\documentclass[runningheads]{llncs}
\usepackage[T1]{fontenc}
\usepackage{graphicx}
\usepackage{color}
\usepackage{makecell}
\usepackage{mathtools}
\usepackage[labelfont=bf]{caption}
\usepackage[labelformat=simple,labelfont=bf]{subcaption}

\usepackage{booktabs,multirow}
\usepackage{array}
\usepackage{pifont}
\usepackage{bbding}
\usepackage[pagebackref=true,breaklinks=true,colorlinks=true,bookmarks=false,citecolor=blue,urlcolor=blue,linkcolor=blue]{hyperref}
\usepackage[misc]{ifsym}
\usepackage{amsmath}
\usepackage{amsfonts}
\usepackage{xspace}
\usepackage{soul}
\usepackage{fancyhdr}
\definecolor{frenchblue}{rgb}{0.0, 0.45, 0.73}

\fancypagestyle{firstpage}{
  \fancyhf{}

  \fancyfoot[L]{* equal contribution}
  \fancyfoot[C]{}
  \fancyfoot[R]{}
}

\def\name{EndoGSLAM\@\xspace}

\makeatother
\begin{document}
\title{EndoGSLAM: Real-Time Dense Reconstruction and Tracking in Endoscopic Surgeries \\ using Gaussian Splatting} 

\titlerunning{EndoGSLAM}
\author{
Kailing Wang*\textsuperscript{1},
Chen Yang*\textsuperscript{1},  
Yuehao Wang\textsuperscript{2},
Sikuang Li\textsuperscript{1},
Yan Wang\textsuperscript{3},
Qi Dou\textsuperscript{2},
Xiaokang Yang\textsuperscript{1},
Wei Shen\textsuperscript{1\dag}
}
\institute{\textsuperscript{1} {MoE Key Lab of Artificial Intelligence, AI Institute, Shanghai Jiao Tong University} \textsuperscript{2} {Dept. of Computer Science and Engineering, The Chinese University of Hong Kong} \textsuperscript{3} {Shanghai Key Laboratory of Multidimensional Information Processing, East China Normal University}}

\maketitle
\thispagestyle{firstpage}
\begin{abstract}
Precise camera tracking, high-fidelity 3D tissue reconstruction, and real-time online visualization are critical for intrabody medical imaging devices such as endoscopes and capsule robots. 
However, existing SLAM (Simultaneous Localization and Mapping) methods often struggle to achieve both complete high-quality surgical field reconstruction and efficient computation, restricting their intraoperative applications among endoscopic surgeries.
In this paper, we introduce EndoGSLAM, an efficient SLAM approach for endoscopic surgeries, which integrates streamlined Gaussian representation and differentiable rasterization to facilitate over 100 fps rendering speed during online camera tracking and tissue reconstructing. 
Extensive experiments show that EndoGSLAM achieves a better trade-off between intraoperative availability and reconstruction quality than traditional or neural SLAM approaches, showing tremendous potential for endoscopic surgeries.
The project page is at \href{https://EndoGSLAM.loping151.com}{https://EndoGSLAM.loping151.com}
\end{abstract}

\section{Introduction}
Endoscopy, a minimally invasive technique for examining and treating internal organs and passages, relies heavily on the skill and precision of operators, especially during complex surgical procedures.
This reliance underscores the vital need for advanced visualization systems that enhance the surgeon's field of view, aid in pinpointing critical areas, and facilitate safer and more efficacious surgical interventions. Key technologies such as endoscopic reconstruction and tracking play a pivotal role in surgical visualization, with Simultaneous Localization and Mapping (SLAM) being a common choice for 
them~\cite{azagra2023endomapper,ali2022we}.

One ideal SLAM approach for surgeries should support online tracking and reconstruction. More importantly, it should enable real-time online visualization of reconstruction, which means it can simultaneously perform tracking, reconstructing, and rendering, allowing surgeons to review any area of interest among previously observed regions at any time. 
Additionally, the method should achieve precise localization and induce complete and high-quality reconstructions.
\begin{figure*}[t]
\centerline{\includegraphics[width=\textwidth]{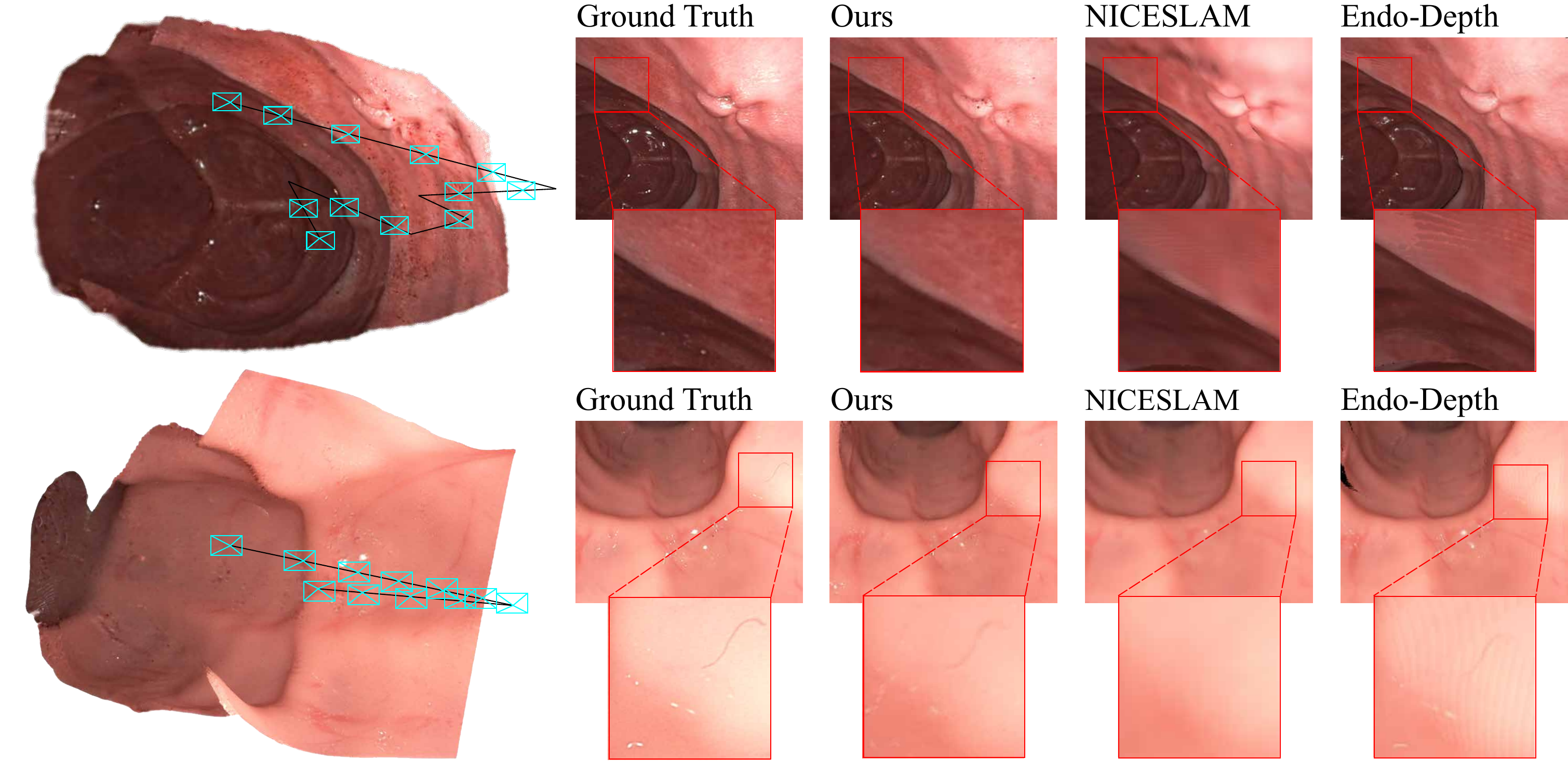}}
\caption{Comparative Visualization of Novel View Synthesis.
From left to right, we show the holistic rendering from \name, the ground truth of one given viewpoint, renderings of \name, NICESLAM~\cite{niceslam} and Endo-Depth~\cite{endodepth}.
These comparisons highlight \name's superior fidelity.}
% \vspace{-1.5 em}
\label{fig: pipeline}
\end{figure*}

Traditional SLAM approaches~\cite{grasa2013visual,mahmoud2017slam,wang2019visual} often yield sparse geometric representations, primarily serving to facilitate endoscope tracking since geometric features are scarce and unreliable among endoscopic procedures~\cite{ozyoruk2021endoslam}. To address this, some approaches~\cite{afsfm,sageslam,ma2021rnnslam,wei2022stereo,posner2023c,endodepth,rau2023bimodal,gu2022vision} have adopted appearance-based optimization for dense mapping and enhanced tracking precision. However, these methods struggle to achieve fine-grained dense reconstructions, impacting novel view rendering and limiting their effectiveness in real-world surgical applications.

Recent advancements in neural rendering, especially Neural Radiance Fields (NeRF)~\cite{nerf} and 3D Gaussian Splatting~\cite{3dgs}, have shown promise for high-fidelity surgical reconstruction~\cite{lerplane,forplane,endonerf}. Several methods~\cite{niceslam,imap,sandstrom2023point,li2023dense,wang2023co,zhu2023nicer} are proposed to integrate NeRF with SLAM. Implicit neural representations, despite offering detailed global maps and photometric capture via differentiable rendering, incur high computational costs, which necessitate pixel sampling for efficiency. This hinders their intraoperative viability in endoscopic contexts.

In this paper, we propose a novel SLAM approach designed for endoscopic surgeries, \name, which simultaneously performs online precise camera tracking, high-quality dense reconstruction, and real-time novel view synthesis.
Specifically, \name designs a simplified Gaussian representation and uses differentiable rasterization to facilitate fast optimization and rendering. Unlike traditional or implicit SLAM representations that depend on sparse geometric features or are limited by inadequate pixel sampling strategies, \name can use dense photometric loss for real-time tracking and reconstruction, making it robust among complex surgical fields. 
Besides, \name iteratively expands 3D Gaussians on those previously unobserved regions and partially refines the reconstructed surgical field, significantly reducing computational costs. 
Extensive evaluations demonstrate \name's advantages in terms of optimization speed, rendering quality, and overall system efficiency, showing its huge potential for advanced surgical navigation.

\begin{figure*}[t]
\centerline{\includegraphics[width=\textwidth]{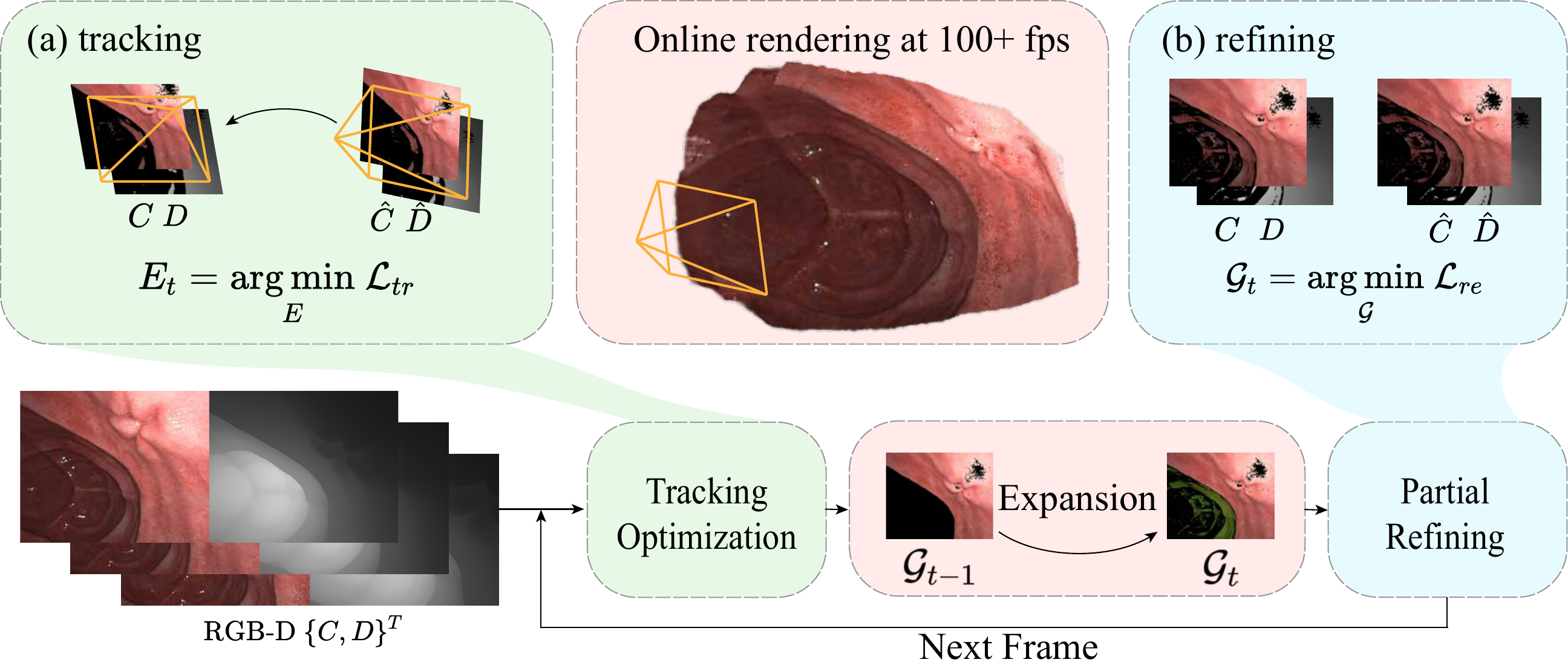}}
\caption{Overview. \name aims to track the camera and reconstruct tissues among endoscopic surgeries while enabling online visualization.}
% \vspace{-1 em}
\label{fig: pipeline}
\end{figure*}

\section{Method}
\name is an efficient dense RGB-D SLAM method for endoscopic procedures utilizing 3D Gaussians as the core representation.
It begins with an innovative modification to the standard 3D Gaussian representation, initializing it to adapt to the complex environments encountered in endoscopy (Sec.~\ref{Method: Preliminaries}).
After the initialization, we leverage differentiable rasterization to enable gradient-based optimization for optimizing the camera pose in each incoming frame (Sec.~\ref{Method: Camera Tracking}). 
We then proceed to expand our 3D Gaussian representation into areas previously unobserved, thus complementing the scene (Sec.~\ref{Method: Gaussian Expanding}).
Finally, we propose a partial refinement strategy for efficiently optimizing the expanded 3D Gaussians (Sec.~\ref{Method: Partial Refinement}). 
The overall framework is illustrated in Fig. \ref{fig: pipeline}.

\subsection{Preliminaries and Initialization} \label{Method: Preliminaries}
To efficiently handle the highly localized illumination characteristic of endoscopic procedures, we propose a streamlined 3D Gaussian representation. 3D Gaussian Splatting~\cite{3dgs} represents complex scenes with collections of 3D Gaussians, each defined by a set of parameters including center location $\mu$, rotation quaternion, scaling vector, opacity $\sigma$, and spherical harmonic (SH) coefficients. 
We first replace SH coefficients with a color attribute $c$ based on the fact that lighting primarily moves with the camera in endoscopy, reducing the need for complex view-dependent effects modeling. 
Besides, we employ a uniform scaling factor for all three dimensions to accelerate optimization. 
In this way, a surgical field is parameterized as a set of isotropic Gaussians:
$\mathcal{G}=\{G_i:\mu_i, c_i, r_i, \sigma_i\}_{i=1}^{N},$ where $r_i$ represents the radius of the $i$-th Gaussian. 
Our simplification significantly reduces the number of parameters to optimize, leading to a significant computational cost reduction of approximately 86\% (59 to 8 parameters). 

We utilize the efficient differentiable 3D Gaussian Splatting algorithm~\cite{3dgs} to render our simplified Gaussian representation. Given a collection of 3D Gaussians $\mathcal{G}$, along with camera pose and intrinsic parameters, our rendering process begins by sorting all Gaussians from near end to far end. Subsequently, we efficiently render an RGB image by alpha-compositing the splatted 2D projection of each Gaussian in the pixel space, determining the color of a pixel $u$ as:
% \vspace{-0.1em}
\begin{align}
\hat C(u) = \sum_{i \in N} c_i \alpha_i \prod_{j=1}^{i-1} (1 - \alpha_j), \quad \alpha_i = \sigma_i \exp\left( -\frac{\| u - \mu_i^{2D} \|^2}{2(r_i^{2D})^2} \right)
\end{align}
where $\mu_i^{2D}$ and $r_i^{2D}$ are the 2D projection of $\mu_i$ and $r_i$, respectively. We estimate the depth $D(u)$ at a pixel $u$ similar to color rendering as the sum of z coordinates of the Gaussians affecting this pixel weighted by the transmittance factor:
% \vspace{-0.1em}
\begin{align}
\hat D(u) = \sum_{i \in N} z_i \alpha_i \prod_{j=1}^{i-1} (1 - \alpha_j),
\end{align}
where $z_i$ is the z coordinate of $\mu_i$. Since $D(u)$ is a weighted sum of $z_i$, we can simply accumulate the weights to represent the visibility of $u$:
\vspace{-0.1em}
\begin{align}
V(u) = \sum_{i \in N} \alpha_i \prod_{j=1}^{i-1} (1 - \alpha_j).
\end{align}
This differentiable rendering process enables us to optimize camera pose and Gaussian parameters via gradient-based optimization.

Initiating from an initial frame, we conduct pixel reprojection into 3D space to construct a point cloud using a known intrinsic matrix and an identity-initialized pose matrix. Subsequently, we convert the point cloud into a set of 3D Gaussians denoted as $\mathcal{G}_{t=0}$. Each point within this Gaussian ensemble is assigned positional coordinates represented by $\mu_i$ and its color converted to $c_i$. The radius $r_i$ is determined as equivalent to a one-pixel radius upon projection into the 2D image, calculated by dividing the depth by the focal length. The opacity parameter $\sigma_i$ is initialized as a constant value (0.5). 

\subsection{Camera Tracking} \label{Method: Camera Tracking}
We employ gradient descent for camera tracking using sequential RGB-D frames. The current pose $\boldsymbol{E}_{t}$ is initialized based on the previous pose $\boldsymbol{E}_{t - 1}$ and constant velocity $\Delta (\boldsymbol{E}_{t - 1}, \boldsymbol{E}_{t - 2})$. We then render the current image $\hat{C}_t$, depth $\hat{D}_t$, and visibility ${V}_t$ using splatting based on $\boldsymbol{E}_{t}$, optimizing the pose $\boldsymbol{E}_{t}$ by minimizing a re-rendering loss. Recognizing that not all pixels contribute equally to accurate tracking, we employ a pre-filter $M_t$ defined on the gray-scale pixel intensities of the current image to exclude pixels with unreliable brightness, \textit{i.e.}, if $\delta \leq G_t(u) \leq 1-\delta$, $M_t(u)=1$; otherwise, $M_t(u)=0$, where $G_t(u)$ is the gray-scale intensity at pixel $u$ in the image $C_t$ and $\delta=0.1$ is a fixed intensity threshold. This approach is necessitated by the unique lighting conditions in the surgical field, where the light source moves with the camera, leading to variability in tissue brightness across frames, and causing insufficient brightness and color in areas further away from the camera. We also utilize the visibility map to identify accurately reconstructed tissues ensuring optimization focuses on these areas, thereby enhancing tracking accuracy. The loss function for camera tracking is:
\begin{align}
\mathcal{L}_{tr} &= \sum_{u} M_t\left(u\right)\cdot V_t(u; \rho_t) \cdot \left(\left|\hat C_t(u) - C_t(u)\right|_1 + \left|\hat D_t(u) - D_t(u)\right|_1\right),
%M(u) &= (\delta \leq G(u) \leq 1-\delta), 
% \cdot (\left|\hat D(u) - D(u)\right|_1 < 20\frac{1}{n}\sum\left|\hat D(u) - D(u)\right|_1),
\end{align}
where ${C}_t(u)$ and ${D}_t(u)$ are the ground truth color and depth of pixel $u$ in the current frame; $V_t(u; \rho_t)$ is an indicator function to indicate whether $V_t(u)$ is greater than a visibility threshold $\rho_t$. $\rho_t$ is fixed to 0.99 through all the experiments. 

\subsection{Gaussian Expanding} \label{Method: Gaussian Expanding}
Following the camera tracking, we update the 3D Gaussian representation to incorporate newly observed tissues. To update the Gaussians $\mathcal{G}_t$, 
the expansion process adheres to three key principles:
1) Areas that fail to represent the current surgical field accurately, 
typically for areas with visibility $V_t(u)$ lower than a visibility threshold $\rho_e$. 
2) Regions identified as containing new geometric details in front of the existing tissue reconstruction surface are also added to $\mathcal{G}_{t}$. % depth mask
3) Pixels that with unreliable color are excluded from expansion. 
Pixels satisfying these criteria are added to $\mathcal{G}_t$ using the same method employed during initialization. This involves reprojection of pixels, and conversion to 3D Gaussians with corresponding color, center location, and other parameters, as detailed in Sec.~\ref{Method: Preliminaries}.

\subsection{Partial Refining} \label{Method: Partial Refinement}
After applying Gaussian expansion, we obtain the updated Gaussians $\mathcal{G}_{t}$. However, newly expanded Gaussians require further optimization for better novel view synthesis. 
We design a partial refining strategy that focuses on those newly expanded Gaussians and recently added sub-optimal Gaussians simultaneously, leading to stable and efficient reconstruction.
Specifically, we designate every k-th frame as a keyframe and then cache them into a keyframe list. To improve efficiency, we assign higher sampling probabilities to keyframes that are temporally or spatially closer to the current frame. 
The probability is derived by: 
% \vspace{-0.1em}
\begin{align}
    P(f_l) = \log_2\left(1 + \frac{1}{d_l + s}\right) + \log_2\left(1 + \frac{1}{t_l + s}\right),
\end{align}
where $f_l$ is the l-th keyframe in the list; $d_l$ and $t_l$ are the L2 distance and time index of $f_l$, scaled down by the distance and time index of the current frame from $0$-th frame; $s$ is a constant to limit the scale of the probability, and is set to 0.2 through all the experiments. We assign a certain probability $p_c$ to the current frame, normalize $P(f_l)$ so that $\sum_{l}P(f_l)=1-p_c$, and utilize this normalized probability distribution to sample keyframes. In each iteration, we select a frame from the list according to its probability and refine $\mathcal{G}_{t}$ using the following loss function:
% \vspace{-0.1em}
\begin{align}
  \mathcal{L}_{re} = \sum_{u} M &\left(u \right) \cdot \Big( (1 - \lambda_{ssim})\left|\hat C(u) - C(u)\right|_1 \nonumber \\
  &\quad + \lambda_{ssim}\left(1-\text{SSIM}(\hat C(u),C(u))\right) + \left|\hat D(u) - D(u)\right|_1 \Big),
\end{align}
where $\hat{C}(u)$, $C(u)$, $\hat{D}(u)$, and $D(u)$ are the rendered color, ground truth color, render depth, and ground truth depth of pixel $u$ in the selected frame, respectively. $\text{SSIM}$ means SSIM loss and $\lambda_{ssim} = 0.2$ across all experiments.

\section{Experiments}

\begin{figure*}[t]
\centerline{\includegraphics[width=\textwidth]{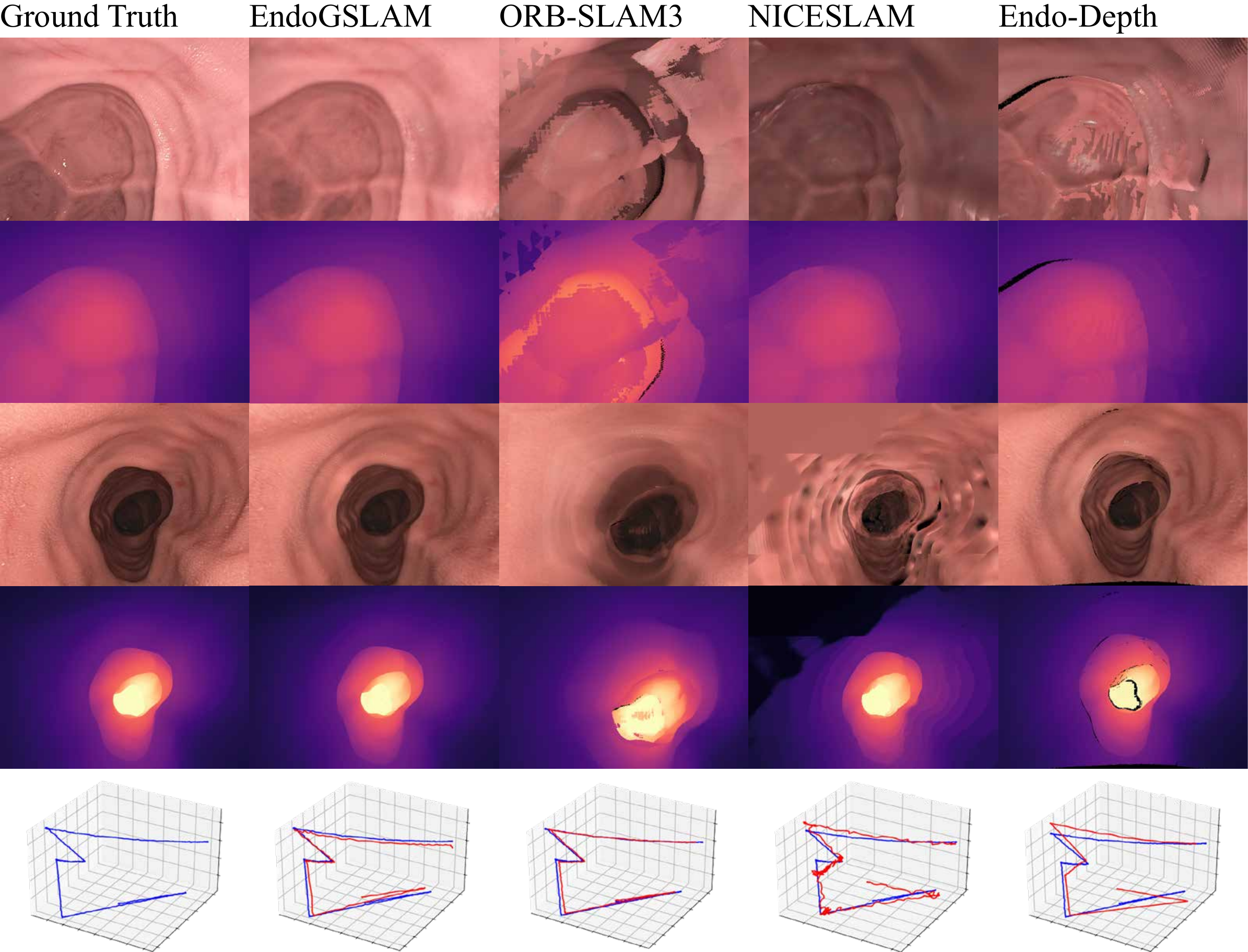}}
\caption{Qualitative results on sequence \textit{cecum\_t2\_b} and \textit{sigmoid\_t2\_a}.}
\vspace{-1.5em}
\label{fig: quant}
\end{figure*}

\subsection{Dataset and Evaluation Metrics}
We evaluate our proposed method on the Colonoscopy 3D Video Dataset (C3VD) \cite{bobrow2023}. This dataset provides ground-truth RGB images, depths, and camera poses for both photometric and geometric evaluation. We choose 10 clips of high-definition clinical colonoscopic videos. Each lasts for 21 seconds and contains 638 frames on average. We pre-undistort the images and the resolution is 675$\times$540. 

For reconstruction, we use the RMSE~\cite{rmse} (mm) on depth for geometric evaluation. As for camera tracking, we use the absolute trajectory (ATE, mm) error to evaluate. We further demonstrate our superior rendering performance using the peak signal-to-noise ratio (PSNR), SSIM~\cite{ssim}, and LPIPS~\cite{lpips}. 

\subsection{Implementation Details} \label{details}
We implement \name mainly with PyTorch~\cite{paszke2017automatic} and CUDA and provide two versions, \textit{i.e.} EndoGSLAM-H (high-quality) and EndoGSLAM-R (real-time). For EndoGSLAM-R, we use $\rho_e = 0.3$ to reproject fewer pixels during expansion, optimize camera poses for 5 iterations/frame at half resolution, and refine for 6 iterations every 2 frames. Keyframes are selected every 4 frames, and we set $p_c = 0.95$ to emphasize the current frame. As for EndoGSLAM-H, we set $\rho_e = 0.5$, optimize camera poses for 15 iterations/frame, and refine for 25 iterations/frame. Keyframes are selected every 8 frames, and $p_c = 0.1$ prioritizes keyframes for quality improvement. All the experiments are done on a machine with Core 13700K CPU and RTX 4090 GPU running Ubuntu 22.04.

\begin{table}[t]
    \setlength\tabcolsep{5pt}
    \centering
    \caption{Quantitative results on the C3VD dataset.}
    \resizebox{\linewidth}{!}{{\begin{tabular}{c|ccccc}
\hline
Methods  & PSNR$\uparrow $        & SSIM$\uparrow$         & LPIPS$\downarrow$         & RMSE(mm)$\downarrow$            & ATE (mm)$\downarrow$    \\ \hline
ORB-SLAM3\cite{orbslam}   & 17.89 {\tiny $\pm$} 2.31   &  0.64 {\tiny $\pm$} 0.10  &  0.35 {\tiny $\pm$} 0.06  &  7.72 {\tiny $\pm$} 2.65  &  0.32 {\tiny $\pm$} 0.09 \\
NICESLAM\cite{niceslam} & 22.07 {\tiny $\pm$} 4.12   & 0.73 {\tiny $\pm$} 0.13  & 0.33 {\tiny $\pm$} 0.07  & 1.88 {\tiny $\pm$} 1.04   & 0.48 {\tiny $\pm$} 0.33  \\
Endo-Depth\cite{endodepth} & 18.13 {\tiny $\pm$} 2.43   & 0.64 {\tiny $\pm$} 0.09 & 0.33 {\tiny $\pm$} 0.06  & 5.10 {\tiny $\pm$} 2.39   &  1.25 {\tiny $\pm$} 0.98 \\ \hline
EndoGSLAM-H & 22.16 {\tiny $\pm$} 2.66   & 0.77 {\tiny $\pm$} 0.08  & 0.22 {\tiny $\pm$} 0.05    &  2.17 {\tiny $\pm$} 1.26     & 0.34 {\tiny $\pm$} 0.21    \\
EndoGSLAM-R & 18.37 {\tiny $\pm$} 2.17   & 0.67 {\tiny $\pm$} 0.10  & 0.30 {\tiny $\pm$} 0.07    & 4.33 {\tiny $\pm$} 2.39    &  1.23 {\tiny $\pm$} 0.90 
      \\ \hline
w.o. Pre-filter & 17.79 {\tiny $\pm$} 2.57   & 0.63 {\tiny $\pm$} 0.14  & 0.32 {\tiny $\pm$} 0.08    & 4.11 {\tiny $\pm$} 2.07    &  2.14 {\tiny $\pm$} 2.33            \\
w.o. Partial Refining & 17.64 {\tiny $\pm$} 2.49   & 0.63 {\tiny $\pm$} 0.12  & 0.31 {\tiny $\pm$} 0.08    & 4.39 {\tiny $\pm$} 2.02    &  1.34 {\tiny $\pm$} 1.13            \\
w.o. Simplification & 17.23  {\tiny $\pm$} 2.45 & 0.65 {\tiny $\pm$} 0.14 & 0.37 {\tiny $\pm$} 0.08 & 4.23 {\tiny $\pm$} 2.42 & 2.26 {\tiny $\pm$} 3.42 \\
 \hline
\end{tabular}
}} %\scriptsize 
    \label{tab: metrics}
    % \vspace{-\baselineskip}
\end{table}

\begin{table}[t]
    \setlength\tabcolsep{5pt}
    \centering
    \caption{Speed on the C3VD dataset.}
    \resizebox{\linewidth}{!}{{\begin{tabular}{c|ccccc}
\hline
\makecell{Methods} & \makecell{tracking\\time/frame} & \makecell{reconstruction\\time/frame} & \makecell{online\\reconstruction} & \makecell{online\\rendering speed} \\ \hline
ORB-SLAM3\cite{orbslam}   & 8.5ms   & 32.3ms  & $\times$ & $\times$ \\
NICESLAM\cite{niceslam} & 140.29ms  & 2558.0ms  & $\checkmark$  &  0.27 fps \\
Endo-Depth\cite{endodepth} & 194.52ms   & 93.7ms  & $\times$ & $\times$   \\
EndoGSLAM-H & 151.4ms   & 268.0ms  & $\checkmark$    & 100+ fps   \\
EndoGSLAM-R & 62.4ms   & 65.1ms  & $\checkmark$    & 100+ fps   \\ \hline
w.o. Simplification & 90.0ms   & 98.0ms  & $\checkmark$    & 100+ fps  \\ \hline
\end{tabular}
}} %\scriptsize 
    \label{tab: time}
    % \vspace{-\baselineskip}
\end{table}

\subsection{Evaluation}
We primarily compare \name to three representative methods:
A well-known traditional SLAM with robust visual tracking and sparse mapping, ORB-SLAM3~\cite{orbslam}; A state-of-the-art dense SLAM based on NeRF~\cite{nerf} that introduces a hierarchical neural implicit representation, NICESLAM~\cite{niceslam}; An endoscopic SLAM that employs photometric constraints to achieve accurate reconstruction and tracking, Endo-Depth~\cite{endodepth}. 
For a fair comparison, all these methods are provided with RGB-D frames.

In Table.~\ref{tab: metrics}, we compare two versions of \name with other methods in terms of novel view rendering, reconstruction, and camera localization performance. 
We also show the average runtime in Tabel \ref{tab: time} and qualitative results in Fig.~\ref{fig: quant}. Only \name achieves online precise tracking, high-quality reconstruction, and real-time online visualization simultaneously, demonstrating its huge potential for intraoperative navigation in endoscopic surgery. Traditional systems, \textit{i.e.} ORB-SLAM3 and Endo-Depth, excel in localization but depend on post-process volumetric fusion for dense reconstruction. This fusion process is sensitive to pose shifts and depth noise, leading to massive fragments in space. 
NICESLAM shows competitive performance but struggles with efficiency, only achieving online rendering speed at 0.27 fps, which is unacceptable for surgeries. Besides, NICESLAM often synthesizes blurred renderings due to its implicit representation. 
In contrast, EndoGSLAM-H utilizes an explicit 3D Gaussian representation to process RGB-D streams at 3 fps and shows better localization, reconstruction, and rendering performance. Moreover, it supports online rendering at over 100 fps, providing robust assistance for surgical procedures. To further support time-sensitive surgical settings, we introduce a real-time variant, EndoGSLAM-R. It prioritizes immediate processing capabilities by making a deliberate trade-off, accepting a slight reduction in performance to achieve real-time process, thus addressing the critical balance between speed and quality necessary for intraoperative assistance.

\subsection{Ablation Study}
We also report our ablation on the pre-filter $M$, the keyframe-based refining strategy and the simplification of Gaussians. in Table.~\ref{tab: metrics}. Metrics are tested on EndoGSALM-R since EndoGSLAM-H is more robust to these variations due to its more training iterations on wider data. Results show that our pre-filter $M$ effectively reduces the influence of unreliable information. Omitting this module leads to artifacts in the reconstruction and instability in the tracking process. 
The keyframe-based refining strategy, which uses previous keyframes to assist training, improves overall performance, particularly in real-time scenarios where efficient training is crucial.
The simplification of Gaussians results in enhanced optimization speeds, as demonstrated in Table~\ref{tab: time}. Additionally, the simplification of SH coefficients contributes to color stability. In the absence of such simplifications, the color becomes contingent upon the viewing direction, leading to pronounced artifacts when observed from a novel view.

\section{Conclusion and Future Work}
In this work, we introduce \name, an advanced dense SLAM framework that enables accurate localization, high-quality reconstruction, and more importantly, online real-time visualization, owing to a streamlined 3D Gaussians representation, differentiable rasterization, and efficient optimization strategy. 
Experiments prove the superior performance of \name compared to traditional and neural SLAM methods, demonstrating its tremendous potential to enhance endoscopic surgical procedures.
Future work aims to eliminate the reliance on depth information, consider minor deformation, and seamlessly integrate it into surgical navigation systems.
\clearpage
{\small
\bibliographystyle{splncs04}
\bibliography{egbib}

\begin{thebibliography}{10}
\providecommand{\url}[1]{\texttt{#1}}
\providecommand{\urlprefix}{URL }
\providecommand{\doi}[1]{https://doi.org/#1}

\bibitem{ali2022we}
Ali, S.: Where do we stand in ai for endoscopic image analysis? deciphering gaps and future directions. npj Digital Medicine  \textbf{5}(1), ~184 (2022)

\bibitem{azagra2023endomapper}
Azagra, P., Sostres, C., Ferr{\'a}ndez, {\'A}., Riazuelo, L., Tomasini, C., Barbed, O.L., Morlana, J., Recasens, D., Batlle, V.M., G{\'o}mez-Rodr{\'\i}guez, J.J., et~al.: Endomapper dataset of complete calibrated endoscopy procedures. Scientific Data  \textbf{10}(1), ~671 (2023)

\bibitem{bobrow2023}
Bobrow, T.L., Golhar, M., Vijayan, R., Akshintala, V.S., Garcia, J.R., Durr, N.J.: Colonoscopy 3d video dataset with paired depth from 2d-3d registration. Medical Image Analysis p. 102956 (2023)

\bibitem{orbslam}
Campos, C., Elvira, R., Rodr{\'\i}guez, J.J.G., Montiel, J.M., Tard{\'o}s, J.D.: Orb-slam3: An accurate open-source library for visual, visual--inertial, and multimap slam. IEEE Transactions on Robotics  \textbf{37}(6),  1874--1890 (2021)

\bibitem{grasa2013visual}
Grasa, O.G., Bernal, E., Casado, S., Gil, I., Montiel, J.: Visual slam for handheld monocular endoscope. IEEE transactions on medical imaging  \textbf{33}(1),  135--146 (2013)

\bibitem{gu2022vision}
Gu, Y., Gu, C., Yang, J., Sun, J., Yang, G.Z.: Vision--kinematics interaction for robotic-assisted bronchoscopy navigation. IEEE Transactions on Medical Imaging  \textbf{41}(12),  3600--3610 (2022)

\bibitem{3dgs}
Kerbl, B., Kopanas, G., Leimk{\"u}hler, T., Drettakis, G.: 3d gaussian splatting for real-time radiance field rendering. TOG  \textbf{42}(4) (2023)

\bibitem{li2023dense}
Li, H., Gu, X., Yuan, W., Yang, L., Dong, Z., Tan, P.: Dense rgb slam with neural implicit maps. arXiv preprint arXiv:2301.08930  (2023)

\bibitem{sageslam}
Liu, X., Li, Z., Ishii, M., Hager, G.D., Taylor, R.H., Unberath, M.: Sage: slam with appearance and geometry prior for endoscopy. In: ICRA. pp. 5587--5593. IEEE (2022)

\bibitem{ma2021rnnslam}
Ma, R., Wang, R., Zhang, Y., Pizer, S., McGill, S.K., Rosenman, J., Frahm, J.M.: Rnnslam: Reconstructing the 3d colon to visualize missing regions during a colonoscopy. Medical image analysis  \textbf{72},  102100 (2021)

\bibitem{mahmoud2017slam}
Mahmoud, N., Hostettler, A., Collins, T., Soler, L., Doignon, C., Montiel, J.M.M.: Slam based quasi dense reconstruction for minimally invasive surgery scenes. arXiv preprint arXiv:1705.09107  (2017)

\bibitem{nerf}
Mildenhall, B., Srinivasan, P.P., Tancik, M., Barron, J.T., Ramamoorthi, R., Ng, R.: Nerf: Representing scenes as neural radiance fields for view synthesis. Communications of the ACM  \textbf{65}(1),  99--106 (2021)

\bibitem{ozyoruk2021endoslam}
Ozyoruk, K.B., Gokceler, G.I., Bobrow, T.L., Coskun, G., Incetan, K., Almalioglu, Y., Mahmood, F., Curto, E., Perdigoto, L., Oliveira, M., et~al.: Endoslam dataset and an unsupervised monocular visual odometry and depth estimation approach for endoscopic videos. Medical image analysis  \textbf{71},  102058 (2021)

\bibitem{paszke2017automatic}
Paszke, A., Gross, S., Chintala, S., Chanan, G., Yang, E., DeVito, Z., Lin, Z., Desmaison, A., Antiga, L., Lerer, A.: Automatic differentiation in pytorch  (2017)

\bibitem{posner2023c}
Posner, E., Zholkover, A., Frank, N., Bouhnik, M.: C 3 fusion: consistent contrastive colon fusion, towards deep slam in colonoscopy. In: International Workshop on Shape in Medical Imaging. pp. 15--34. Springer (2023)

\bibitem{rau2023bimodal}
Rau, A., Bhattarai, B., Agapito, L., Stoyanov, D.: Bimodal camera pose prediction for endoscopy. IEEE Transactions on Medical Robotics and Bionics  (2023)

\bibitem{endodepth}
Recasens, D., Lamarca, J., F{\'a}cil, J.M., Montiel, J., Civera, J.: Endo-depth-and-motion: Reconstruction and tracking in endoscopic videos using depth networks and photometric constraints. RAL  \textbf{6}(4),  7225--7232 (2021)

\bibitem{sandstrom2023point}
Sandstr{\"o}m, E., Li, Y., Van~Gool, L., Oswald, M.R.: Point-slam: Dense neural point cloud-based slam. In: Proceedings of the IEEE/CVF International Conference on Computer Vision. pp. 18433--18444 (2023)

\bibitem{afsfm}
Shao, S., Pei, Z., Chen, W., Zhu, W., Wu, X., Sun, D., Zhang, B.: Self-supervised monocular depth and ego-motion estimation in endoscopy: Appearance flow to the rescue. Medical image analysis  \textbf{77},  102338 (2022)

\bibitem{rmse}
Sturm, J., Engelhard, N., Endres, F., Burgard, W., Cremers, D.: A benchmark for the evaluation of rgb-d slam systems. In: 2012 IEEE/RSJ international conference on intelligent robots and systems. pp. 573--580. IEEE (2012)

\bibitem{imap}
Sucar, E., Liu, S., Ortiz, J., Davison, A.J.: imap: Implicit mapping and positioning in real-time. In: Proceedings of the IEEE/CVF International Conference on Computer Vision. pp. 6229--6238 (2021)

\bibitem{wang2019visual}
Wang, C., Oda, M., Hayashi, Y., Kitasaka, T., Honma, H., Takabatake, H., Mori, M., Natori, H., Mori, K.: Visual slam for bronchoscope tracking and bronchus reconstruction in bronchoscopic navigation. In: Medical Imaging 2019. vol. 10951, pp. 51--57. SPIE (2019)

\bibitem{wang2023co}
Wang, H., Wang, J., Agapito, L.: Co-slam: Joint coordinate and sparse parametric encodings for neural real-time slam. In: CVPR. pp. 13293--13302 (2023)

\bibitem{endonerf}
Wang, Y., Long, Y., Fan, S.H., Dou, Q.: Neural rendering for stereo 3d reconstruction of deformable tissues in robotic surgery. In: MICCAI. pp. 431--441. Springer (2022)

\bibitem{ssim}
Wang, Z., Bovik, A.C., Sheikh, H.R., Simoncelli, E.P.: Image quality assessment: from error visibility to structural similarity. IEEE transactions on image processing  \textbf{13}(4),  600--612 (2004)

\bibitem{wei2022stereo}
Wei, R., Li, B., Mo, H., Lu, B., Long, Y., Yang, B., Dou, Q., Liu, Y., Sun, D.: Stereo dense scene reconstruction and accurate localization for learning-based navigation of laparoscope in minimally invasive surgery. IEEE Transactions on Biomedical Engineering  \textbf{70}(2),  488--500 (2022)

\bibitem{forplane}
Yang, C., Wang, K., Wang, Y., Dou, Q., Yang, X., Shen, W.: Efficient deformable tissue reconstruction via orthogonal neural plane. arXiv preprint arXiv:2312.15253  (2023)

\bibitem{lerplane}
Yang, C., Wang, K., Wang, Y., Yang, X., Shen, W.: Neural lerplane representations for fast 4d reconstruction of deformable tissues. arXiv preprint arXiv:2305.19906  (2023)

\bibitem{lpips}
Zhang, R., Isola, P., Efros, A.A., Shechtman, E., Wang, O.: The unreasonable effectiveness of deep features as a perceptual metric. In: Proceedings of the IEEE conference on computer vision and pattern recognition. pp. 586--595 (2018)

\bibitem{zhu2023nicer}
Zhu, Z., Peng, S., Larsson, V., Cui, Z., Oswald, M.R., Geiger, A., Pollefeys, M.: Nicer-slam: Neural implicit scene encoding for rgb slam. arXiv preprint arXiv:2302.03594  (2023)

\bibitem{niceslam}
Zhu, Z., Peng, S., Larsson, V., Xu, W., Bao, H., Cui, Z., Oswald, M.R., Pollefeys, M.: Nice-slam: Neural implicit scalable encoding for slam. In: CVPR. pp. 12786--12796 (2022)

\end{thebibliography}
}
\end{document}